# Number of Clusters in a Dataset: A Regularized K-means Approach

**Behzad Kamgar-Parsi**
Mathematics, Computer and Information Sciences Division
Office of Naval Research
Arlington, VA 22203

**Behrooz Kamgar-Parsi**
Information Technology Division
Naval Research Laboratory
Washington, DC 20375

## Abstract

Finding the number of meaningful clusters in an unlabeled dataset is important in many applications. Regularized k-means algorithm is a possible approach frequently used to find the correct number of distinct clusters in datasets. The most common formulation of the regularization function is the additive linear term $\lambda k$, where $k$ is the number of clusters and $\lambda$ a positive coefficient. Currently, there are no principled guidelines for setting a value for the critical hyperparameter $\lambda$. In this paper, we derive rigorous bounds for $\lambda$ assuming clusters are *ideal*. Ideal clusters (defined as $d$-dimensional spheres with identical radii) are close proxies for k-means clusters ($d$-dimensional spherically symmetric distributions with identical standard deviations). Experiments show that the k-means algorithm with additive regularizer often yields multiple solutions. Thus, we also analyze k-means algorithm with multiplicative regularizer. The consensus among k-means solutions with additive and multiplicative regularizations reduces the ambiguity of multiple solutions in certain cases. We also present selected experiments that demonstrate performance of the regularized k-means algorithms as clusters deviate from the ideal assumption.

## 1 Introduction

Finding the correct number of distinct and meaningful clusters in an unlabeled dataset is a long-standing and important problem in many applications. These include finding the number of genes in an expression profile, number of distinct cells in cancer research, number of distinct textures in an image, number of speakers in a crowded room from audio recordings, number of vessels in a high-traffic channel from navigation radar signals, and many others. This is an illposed problem and in some cases may not have a unique or clear solution, particularly when the data distributions belonging to clusters are not compact or datasets are contaminated by noise. In this paper we investigate the utility of regularized k-means (also called penalized k-means) algorithms for determining the number of distinct clusters in a dataset.

A widely used approach for this purpose is the k-means algorithm, and while it is appropriate for *same-size spherically symmetric clusters* in Euclidean spaces, in practice it is often used beyond its valid range for clusters with significantly different shapes and sizes. Because of its applicability in many areas including statistics, signal processing, information theory, data compression, image and pattern analysis and others, there are a number of excellent treatments of k-means in the literature. Here we cite only a few [7][10][16][18][27][29][30][38]. For the sake of completeness, we include a brief description of the k-means here as well.

The problem may be stated as: Given a set of $N$ unlabeled data points $\{\mathbf{x}_i\}$, $i = 1, \ldots, N$, in a $d$-dimensional feature space, with a given metric distance (usually Euclidean), partition the data into $k$ clusters $\{\mathcal{C}_j\}$, $j = 1, \ldots, k$, such that the empirical error $E$ (or cost function, distortion, sum of within-cluster variances),

$$E_k = \sum_{j=1}^{k} \sum_{\mathbf{x}_i \in \mathcal{C}_j} \| \mathbf{x}_i - \mathbf{c}_j \|^2, \text{ with } \mathbf{c}_j = \frac{1}{N_j} \sum_{\mathbf{x}_i \in \mathcal{C}_j} \mathbf{x}_i, \tag{1}$$

is minimized. Here $\mathbf{c}_j$ denotes the centroid of cluster $\mathcal{C}_j$, $N_j$ is the number of data points that are in cluster $\mathcal{C}_j$, and $N = \sum_{j=1}^{k} N_j$. The k-means algorithm is a popular method for solving this non-convex optimization problem.

In the k-means algorithm we start by randomly selecting $k$ cluster centroids, and then use the Lloyd iterations to obtain a solution. The Lloyd iteration [28] is a simple Expectation-Maximization (EM) process, more precisely a hard EM process [30], where the two steps are: (i) assign the points that are closest to a cluster centroid to that cluster, (ii) recompute the cluster centroids. Then repeat the two steps until cluster memberships no longer change. This process monotonically decreases the error and typically converges to a local minimum. In order to potentially find a better solution, the algorithm is restarted many times with different initial cluster centroids and then the clustering with the smallest error is taken as the solution. Various initialization procedures have been introduced in the literature among which we cite the k-means++ [4], and its modifications [5][6], that generally yield better solutions. For a comprehensive review and discussion of k-means initializations see the recent paper [13].

Also, a number of alternative algorithms have been proposed in the literature for more efficient solutions. For a comprehensive discussion we refer to [19]. Here we cite only a few: [23] computes distances more efficiently; [26] presents a parallel method; [31] proposes successive partitioning of data; and [20][21] propose a soft k-means algorithm based on the Hopfield network that generally leads to better solutions, albeit at higher computational cost. There are also a body of approximate solutions for k-means; see [1] and references therein.

We note that these algorithms, even for the simpler case of *ideal* clusters (defined below in Sec. 3), cannot be proven to find the best solution, or yield objective information about the correct number of clusters. Neither are the results reproducible in general, because of the random selection of initial clusters.

As noted above, in many applications we do not know the number of clusters a priori, rather we want to find the correct number of clusters. Since increasing $k$ leads to clusters with ever decreasing $E$, the error (1) cannot indicate what the correct $k$ may be. To deal with this problem, a number of cluster analysis methods have been proposed in the literature, which we briefly discuss in Sec. 2. An alternative is to use a regularized error function that has its global minimum at the correct value of $k$. Regularized, or penalized[1], error may be formulated with either *additive* or *multiplicative* regularization functions respectively given by:

$$E_k^{(a)} = E_k + \lambda f(k), \tag{2}$$

$$E_k^{(m)} = f(k)\, E_k, \tag{3}$$

where $E$ is the error given in (1) and $f(k)$ is a monotonically increasing function of $k$. The common practice is to use the additive regularized error function $E^{(a)}$ with linear penalty $\lambda k$. It will have a global minimum at some finite $k$, which depends on the value of the penalty coefficient $\lambda$. The problem then becomes the more complicated simultaneous determination of the optimal $k$, as well as the optimal set of cluster centroids $\{\mathbf{c}_j\}$, $j = 1, \ldots, k$. The correct number of clusters may depend critically on the value of coefficient $\lambda$. Obviously, small $\lambda$ favors a larger number of clusters and large $\lambda$ leads to a smaller number of clusters.

In this paper, we present a rigorous analysis of the regularized k-means with additive penalty to obtain principled bounds for the value of $\lambda$. Since treating general clusters appears to be intractable, we consider ideal clusters that are the simplest form of clusters. We describe and justify the assumption for using ideal clusters in Sec. 3. But briefly, ideal clusters are $d$-dimensional spheres, with identical sizes, with no overlaps, and no outlier background data points.

Experiments show that when clusters deviate from the ideal assumption, the regularized k-means algorithm with additive penalty often yields multiple solutions. Hence, we also examine the regularized k-means algorithm with multiplicative penalty. The consensus among the two often resolves the ambiguity of multiple solutions.

## 2  Related Work

The goal of this paper is to investigate the utility of regularized k-means and provide a more principled basis for it, rather than a comparison of various techniques. Nevertheless, we include a brief discussion of related work here. Finding the number of clusters in a dataset is an old problem for which as yet there are no entirely satisfactory approaches. While a number of methods have been proposed for the un-regularized standard k-means (1) (for example, see the Wikipedia page [39]), there is little work on the regularized k-means formulations. Determining the number of clusters in the standard k-means approaches are based on detecting a noticeable change, or jump or gap, in a defined signature. While in the regularized k-means approaches we seek the minimum of the clustering errors (2) or (3) as a function of $k$. In either approach, the data is first clustered for a large range of $k$ values and then analyzed to estimate which value of $k$ best satisfies the appropriate criterion.

---

[1]We use regularized and penalized interchangeably.

For the non-penalized k-means, we cite only a few relevant papers and the references therein. These approaches include the heuristic elbow method, which looks for a noticeable elbow in the plot of $E_k$ as a function of $k$. The gap statistics method [37] provides a theoretical formalism for the elbow method. In the gap statistics method the pooled sum of within-cluster pairwise distances for each $k$ is compared to the distance in a reference data distribution (typically the null distribution, $k = 1$). The largest gap between these two distances indicates the number of clusters. In particular, this method can show if distinct clusters are not present in the data.

The silhouette method [32] is based on the similarity of within-cluster data points compared to the similarity with the data in the closest neighboring cluster. The $k$ for which the difference between these two similarities is the largest is estimated to be the best $k$. The slope statistic method [15] is an extension of the silhouette method.

In the jump method [35] the average within-cluster distance, $D(k)$, for each $k$ ranging from 1 to a large value is computed. Then where the $k$ for which there is a significant jump in the $D(k)$ vs. $k$ curve is taken as the best $k$. The methods mentioned above are rather subjective.

In the cross-validation method the data is divided into training and validation sets [2][14]. This is a more reliable approach, but the outcome can be sensitive to how the data is divided into training and validation sets.

Another approach is based on an information criterion such as Bayesian or Akaike, which may work well if the data distribution can be represented by a Gaussian mixture model [8] [34].

For the additive regularized k-means, in [25] they derive the k-means with linear additive penalty from a Bayesian nonparametric viewpoint. However, coefficient $\lambda$ is selected as the largest distance any data point can have from its centroid. Although this scheme limits the growth of number of clusters, it does not find the correct number of clusters. In [11] they use an additive penalty term $f(k) = \lambda \sqrt{k/n}$ and determine $\lambda$ empirically from the data through the use of slope heuristics. In our approach, the bounds for $\lambda$ can be derived rigorously; see Appendix C.

For the multiplicative penalty case, we cite [24] where they propose the penalty function, $f(k) = k^{2/d}$, based on the assumption that clusters have uniform rectangular distributions, and argue maximizing the ratio below yields the optimum $k$,

$$(E_{k-1}^{(m)} - E_k^{(m)})/(E_k^{(m)} - E_{k+1}^{(m)}), \quad \text{where} \quad E_k^{(m)} = k^{2/d} E_k.$$

However, as the dimension $d$ increases this penalty function becomes weaker and $E_k^{(m)} \to E_k$, which reduces to the regular k-means.

We show that the parameter-free formulation, $E_k^{(m)}$, has properties that appear to be more useful than the additive penalized error. We also note that the regularized k-means is in the broad class of slope-based methods, but without being subjective in principle.

In the following sections, we define ideal clusters, analyze regularized k-means with additive and multiplicative regularizers, and present a number of selected experiments that show the validity of theoretical analyses and certain limits of ideal cluster assumption.

## 3 Ideal Clusters

Here we define ideal clusters and argue that they are consistent with the underlying assumptions for k-means clusters even though more restrictive. The original k-means clustering may be interpreted as: we want to simultaneously estimate the value of $k$ entities $\{\mathbf{c}_j\}$, from $N$ unlabeled measurements $\{\mathbf{x}_i\}$, and that these measurements have identical independent normal distributions $\mathcal{N}(\mathbf{c}_j, \sigma^2 I)$, with the same covariance $\sigma^2$, around their true values. To do so, we must cluster the measurements into correct groups and maximize the likelihood,

$$P = \frac{1}{Z} \prod_{j=1}^{k} \prod_{\mathbf{x}_i \in \mathcal{C}_j} \exp(-\frac{\|\mathbf{x}_i - \mathbf{c}_j\|^2}{2\sigma^2}), \tag{4}$$

or equivalently minimize the negative of its logarithm $E_k$ given by (1). Here $Z = (2\pi\sigma^2)^{\frac{dN}{2}}$ is the normalization factor, and $d$ is the dimension of the feature space. The underlying assumptions are: (a) clusters are spherically symmetric since the variances are isotropic in all the Euclidean space dimensions, (b) clusters have the same size since variances are identical, (c) clusters are sufficiently separated so no two clusters overlap in such a way that their peaks cannot be distinguished from each other, and (d) there are no outlier data points not belonging to any of the clusters. We should note that the assumption of normal distribution for k-means clusters is not necessary, rather other distributions with spherical symmetry and the same variance are valid.

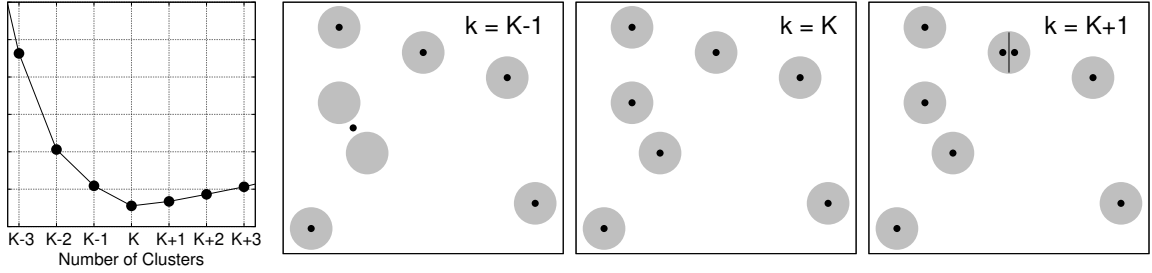

Figure 1: Sketch of a regularized error with its global minimum at the correct number of clusters $K$ (left). A 2D example with $K = 7$ ideal clusters, illustrating results of the optimal algorithm for $k = K - 1$ with 5 spheres and 1 dumbbell (middle left), for $k = K$ with 7 spheres (middle right), and for $k = K + 1$ with 6 spheres and 2 half-spheres (right). We denote the correct number of clusters by upper-case $K$, while lower-case $k$ is a variable for the number of clusters.

We define *ideal clusters* to have the following properties: (a) clusters are spherical; (b) have the same size and volume $V$; (c) are non-overlapping, a stronger assumption than the original; and (d) have no outliers. We make an additional assumption (e) that clusters are dense, that is, the data points nearly fill the volume of the spheres. This is primarily for computational convenience to replace sums with integrals, and implies the number of data points in each cluster may be approximated by the volume $V$ of the sphere and that $N_j$ is equivalent to $V$.

In standard k-means we know the number of clusters a priori. In regularized k-means, however, $k$ is a variable to be determined along with the set of cluster centroids $\{\mathbf{c}_j\}$. The optimization procedure is to vary $k$ and then for each $k$ find the optimal set $\{\mathbf{c}_j\}$ using the standard k-means clustering. The expectation is that $E_k^{(a)}$ as a function of the number of clusters behaves as shown in Fig. 1, where the correct $k = K$ is the value at which the regularized error is minimum. (We denote the correct number of clusters by upper-case $K$.) This happens if the standard k-means algorithm finds the optimal set $\{\mathbf{c}_j\}$ for each $k$. Since the standard k-means algorithm cannot guarantee optimality, we restrict the discussion to ideal clusters for which we can design algorithms that are provably optimal. Below, we present two such algorithms. The initializtion procedure in these two algorithms ensures that when $k$ is equal to the correct number of clusters, then one and only one data point in each cluster is selected as the initial centroid, which in turn ensures that the algorithm converges to the correct and optimal centroids.

### 3.1 Algorithm 1

- Step 1: Choose a data point and save it as the initial centroid for the first cluster $\mathbf{c}_1^0$. (The algorithm is insensitive to the initial choice; we typically choose the point closest to the origin of the feature space.)
- Step 2: Choose the data point farthest from $\mathbf{c}_1^0$. Save it as the initial centroid for the second cluster $\mathbf{c}_2^0$. Using these initial centroids run the Lloyd iteration until convergence to $\mathbf{c}_j$, $j = 1, 2$.
- Step 3: Choose the data point farthest from both $\mathbf{c}_1^0$ and $\mathbf{c}_2^0$. Save it as the initial centroid for the third cluster $\mathbf{c}_3^0$. Using these initial centroids run the Lloyd iteration until convergence to $\mathbf{c}_j$, $j = 1, 2, 3$.
- Step 4 and higher: At step $M$ continue choosing the farthest point to all the previous $M - 1$ initial centroids and save it as the initial centroid for the $M$-th cluster. Using these initial centroids run the Lloyd iteration until convergence to $\mathbf{c}_j$, $j = 1, \ldots, M$.

We note that this initialization procedure may be considered as a deterministic version of k-means++ initialization [4].

When $k$ equals the correct number of clusters, $k = K$, the clustering is optimal.[2] The proof is straightforward, because each initial centroid is in a distinct cluster, and since clusters are non-overlapping each initial centroid converges to the true centroid of the cluster to which it belongs. When $k = K + 1$ one of the clusters splits into two equal half-spheres, since that cluster contains two of the initial clusters.

---

[2]We denote the correct number of clusters by upper case $K$, and the number of clusters as a variable by the lower case $k$.

When $k = K - 1$ cluster configurations become more complex. Most often two adjacent clusters merge and form a dumbbell shaped cluster. Fig. 1 illustrates these cases for an example with $K = 7$. It could also happen, especially when some clusters are close to each other that three clusters are shared by two centroids forming partial or uneven dumbbells sketched in Fig. 14. We will analyze both cases of perfect and uneven dumbbells in Appendix B and show that perfect dumbbells yield tighter upper bound in most cases.

### 3.2 Algorithm 2

- Step 1: Choose a data point and save it as the initial centroid for the first cluster $\mathbf{c}_1^{(1)}$. (The algorithm is insensitive to the initial choice; we typically choose the point closest to the centroid of all the points.)

- Step 2: Set $k = 2$. Choose the data point farthest from $\mathbf{c}_1^{(1)}$. Save it as the initial centroid for the second cluster $\mathbf{c}_2^{(1)}$. Using $\mathbf{c}_j^{(1)}$, $j = 1, 2$, as the initial centroids run the Lloyd iteration until convergence to $\mathbf{c}_j^{(2)}$, $j = 1, 2$.

- Step 3: Set $k = 3$. Choose the data point farthest from both $\mathbf{c}_1^{(2)}$ and $\mathbf{c}_2^{(2)}$. Save it as the initial centroid for the third cluster $\mathbf{c}_3^{(2)}$. Using $\mathbf{c}_j^{(2)}$, $j = 1, 2, 3$, as initial centroids run the Lloyd iteration until convergence to $\mathbf{c}_j^{(3)}$, $j = 1, 2, 3$.

- Step 4 and higher: At step $M$ set $k = M$ and continue choosing the farthest point to all the previous $M - 1$ centroids, $\mathbf{c}_j^{(M-1)}$, $j = 1, \ldots, M - 1$, and save it as the initial centroid for the $M$-th cluster $\mathbf{c}_M^{(M-1)}$. Using $\mathbf{c}_j^{(M-1)}$, $j = 1, \ldots, M$ as the initial centroids run the Lloyd iteration until convergence to $\mathbf{c}_j^{(M)}$, $j = 1, \ldots, M$.

### 3.3 Errors for Individual Clusters

For the analysis of k-means with additive and multiplicative regularizations we need errors for individual ideal clusters that take on three different shapes, namely, sphere with clustering error $E_s$, half sphere with error $E_h$, and perfect dumbbell with error $E_d$. These individual clustering errors are given by,

$$
\begin{cases}
E_s & = V R^2 \alpha \\
E_h & = V R^2 \beta \\
E_d & = 2 E_s + \frac{1}{2} V L^2 = 2 V R^2 \alpha + \frac{1}{2} V L^2
\end{cases}
\tag{5}
$$

where $V$ is the volume of the $d$-dimensional sphere with radius $R$,

$$
V = \frac{\pi^{\frac{d}{2}}}{\Gamma(\frac{d+2}{2})} R^d,
\tag{6}
$$

$L$ is the distance between the centers of the two spheres forming the perfect dumbbell,

$$
\alpha = \frac{d}{d+2}, \ \beta = \frac{1}{2}\left(\alpha - \gamma^2\right), \ \gamma = \frac{\Gamma(\frac{d+2}{2})}{\sqrt{\pi}\,\Gamma(\frac{d+3}{2})}, \ \rho = R\gamma.
\tag{7}
$$

$\Gamma(\cdot)$ is the generalized factorial function, and $\rho$ is the distance of the half-sphere centroid from its equator plane. We refer to Appendix A for derivations of (5) and (6). Also see [7] for statistics in high-dimensional spaces.

## 4 Additive Regularization

Here we analyze the linear additive regularizer $f(k) = k$ to obtain the bounds for the penalty coefficient $\lambda$. Other functional forms of regularizer are discussed in more detail in Appendix C and mentioned briefly at the end of this section. The conclusions, however, remain the same.

The lower bound for $\lambda$ is uniquely derived from $k = K$ and $k = K + 1$ because ideal clusters are either spheres or half-spheres. The upper bound is derived from $k = K - 1$ and $k = K$. However, when $k = K - 1$ clusters may come in different shapes, such as perfect dumbbells or partial dumbbells. In Appendix B we analyze the different cases and show that the perfect dumbbell is more general and yields the tightest bound for all dimensions. Moreover, our analysis shows that the upper bound obtained from any of these cases are similar and close. Below, we present the case when one of the clusters is a perfect dumbbell.

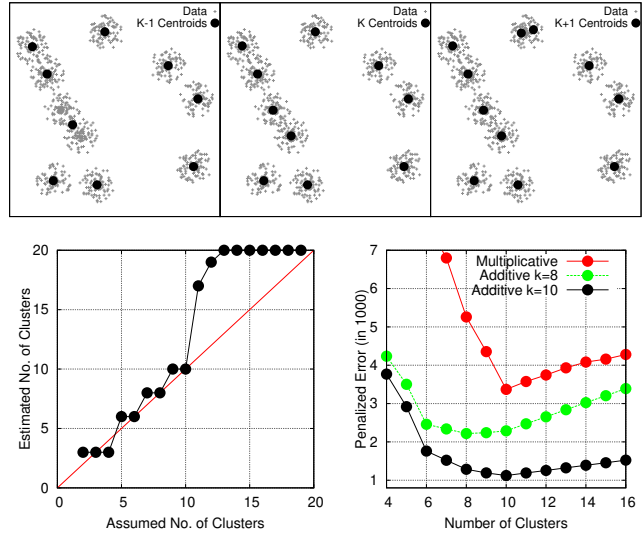

Figure 2: A 2D examples with 10 clusters, each cluster with approximately 100 points. **Top row:** Left figure is when $k = 9$, where one of the clusters is a dumbbell. Middle figure is when $k = 10$, which shows the correct number of clusters. Right figure is when $k = 11$, where one of the clusters splits in half. **Bottom row:** Left figure shows that the additive penalty yields $k = 3, 6, 8, 10$ as solution candidates. Right figure shows multiplicative penalized error yielding $k = 10$ as the only possible solution. We also plot additive penalized errors for $\lambda$ obtained with $k = 8, 10$. Also note that the penalized errors with additive penalty have shallower minima.

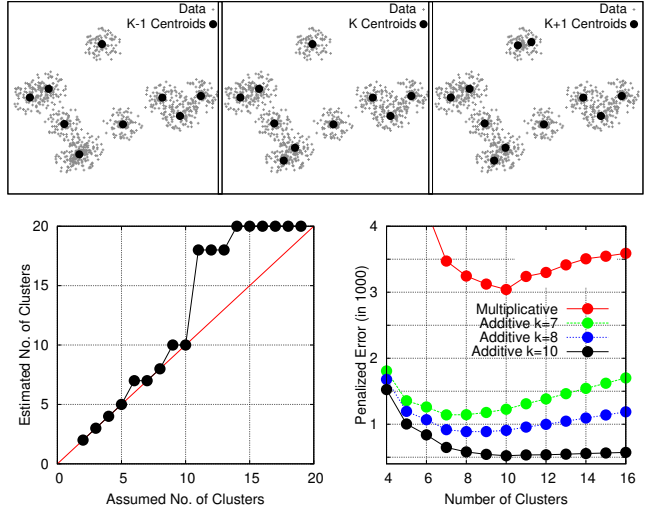

Figure 3: This example is similar to Fig. 2. Clusters are the same but their locations are changed to create more overlaps, which means they are further from the ideal cluster assumption than those in Fig. 2. Additive penalty yields $k = 2, 3, 4, 5, 7, 8, 10$ are solution candidates, while the multiplicative penalty shows $k = 10$ as the only solution.

Let $K$ denote the correct number of spherical (ideal) clusters. The case with $K - 1$ clusters then has $K - 2$ spheres and one dumbbell, and the case with $K + 1$ clusters has $K - 1$ spheres and 2 half-spheres. Thus, the clustering errors for the non-penalized case $E$ in (1) are given by:

$$
\begin{cases}
E_{K-1} = (K - 2)E_s + E_d = K V R^2 \alpha + \frac{1}{2} V L^2 \\[2mm]
E_K \quad = K E_s = K V R^2 \alpha \\[2mm]
E_{K+1} = (K - 1)E_s + 2E_h = (K - 1)V R^2 \alpha + 2 V R^2 \beta.
\end{cases}
\tag{8}
$$

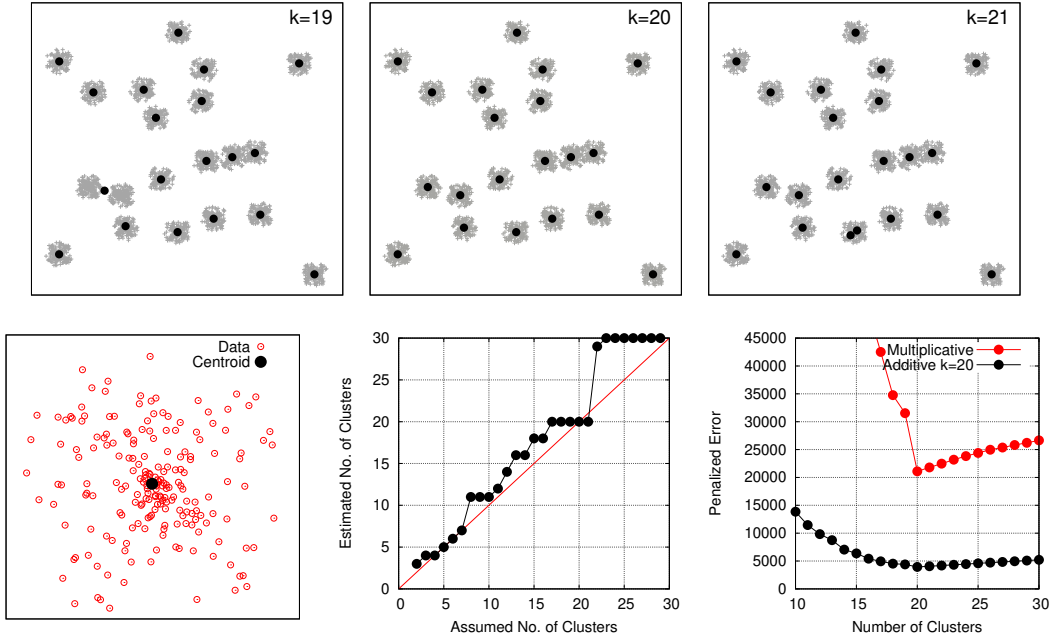

Figure 4: A 2D example with 20 clusters, each cluster with approximately 200 points. **Top row** figures show the result of clustering with $k = 19, 20, 21$. Note that when $k = 19$ one of the clusters is a dumbbell, and when $k = 21$ one of the clusters splits in half. **Bottom row** left shows distribution of data points in a typical cluster. Middle figure shows *Estimated* vs. *Assumed* number of clusters for the additive regularizer with $K = 4, 5, 6, 7, 20$ as candidate solutions. Right figure shows that $E_k^{(m)}$ has a solution at $K = 20$ only. Also plotted is $E_k^{(a)}$, which has a shallow minimum. "Additive $k = 20$" means the curve is computed with $\lambda$ obtained from $K = 20$. These results are obtained by Algorithm 1.

The clustering error obviously decreases as we increase the number of clusters, and it is straight forward to show this for the ideal case:

$$
\begin{cases}
\delta_{K-1,K} = & E_{K-1} - E_K = \frac{1}{2} V L^2 > 0, \\
& \text{for all } d \geq 1 \text{ and } K > 1, \\
\delta_{K,K+1} = & E_K - E_{K+1} = V R^2 (\alpha - 2\beta) > 0, \\
& \text{for all } d \geq 1 \text{ and } K \geq 1.
\end{cases}
\tag{9}
$$

The second inequality is satisfied since $\alpha/2\beta > 1$, as may also be seen in Fig. 8.

For the clustering error $E^{(a)}$ to have the minimum (or a local minimum) at $K$ we must require:

$$
\begin{cases}
\Delta_{K-1,K}^{(a)} = & E_{K-1} - E_K = \frac{1}{2} V L^2 - \lambda > 0, \\
& \text{for all } d \geq 1 \text{ and } K > 1, \\
\Delta_{K,K+1}^{(a)} = & E_K - E_{K+1} = V R^2 (\alpha - 2\beta) - \lambda < 0, \\
& \text{for all } d \geq 1 \text{ and } K \geq 1.
\end{cases}
\tag{10}
$$

Since $R^2(\alpha - 2\beta) = R^2 \gamma^2 = \rho^2$, and volume $V$ stands for the average number of points in each cluster, i.e. $V \approx N/K$, the two inequalities may be combined and written as:

$$
\frac{N\rho^2}{K} < \lambda < \frac{NL^2}{2K}.
\tag{11}
$$

Note that $\lambda$ depends on the correct number of clusters $K$, the hyper-parameter we want to determine. This introduces another layer of complexity in using k-means with additive penalty. To overcome this problem, we propose the following procedure. Also, note that (11) suggests the penalty term, $\lambda K$, does not depend on the number of correct clusters.

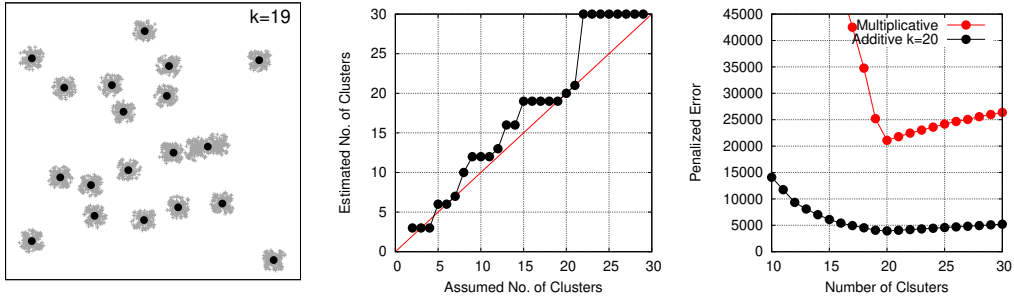

Figure 5: The clusters shonw on the left are the same as in Fig. 4, but the results are obtained by using Algorithm 2. The results for $k = 20, 21$ (and any $k \geq 20$) are identical to those obtained by Algorithm 1 shown in Fig. 4. However, clustering results for $k = 19$ (or any $k < 20$), are different. Note that the dumbbell-shaped cluster here is different from that in Fig. 4 with $k = 19$. The middle figure shows *Estimated* vs. *Assumed* number of clusters for the additive penalty, with the results that $K = 3, 6, 7, 19, 20, 21$ are solution candidates. The right figure shows that $E_k^{(m)}$ has a solution at $K = 20$ only. Also plotted is $E_k^{(a)}$ for $\lambda$ obtained with $K = 20$, which has a shallow minimum.

### 4.1 Procedure for Using Additive Regularization

- Begin by assuming $K = 2$. Select a value for $\lambda$ in the range given by (11). Run the k-means algorithm for $k = 2, \ldots M$, where $M$ is very large. If $E^{(a)}$ has a minimum at $k = 2$, then the assumed $K = 2$ is a potential solution.

- Next assume $K = 3$. Select a value for $\lambda$ in the corresponding range. Run the k-means algorithm for $k = 2, \ldots M$. If $E^{(a)}$ has a minimum at $k = 3$, then the assumed $K = 3$ is a potential solution.

- Then assume $K = 4$, and repeat the process. If $E^{(a)}$ has a minimum at $k = 4$, then $K = 4$ is a potential solution.

- And so on assuming $K = 5, 6, \ldots$ up to a very large value.

To make this procedure consistent and repeatable, we propose to select the midpoint of the range given in (11) as the value for $\lambda$. That is,

$$\lambda = \frac{N(\rho^2 + \frac{1}{2}L^2)}{2K} \approx \frac{NL^2}{4K}.$$ (12)

The approximation is reasonable since (a) $\rho = R\gamma$ and $\gamma^2 = 0.18$ for $d = 2$ and goes to zero as the dimension $d \to \infty$, and (b) the $\rho^2$ term may be neglected since $L \geq 2R$ for non-overlapping clusters. We use the approximate value for $\lambda$ in the experiments we present later. This approximation becomes increasingly better as dimension $d$ increases. For $L$ we choose the smallest inter-centroid distance, which also depends on $K$.

We also considered other penalty functions, namely, $f(k) = \ln k, k^p, e^k$. Similar to the linear penalty (12), $\lambda$ for these cases also depends on $K$ and is, respectively (see Appendix C),

$$\lambda \propto NL^2(1 + \frac{1}{K}), \quad \lambda \propto \frac{NL^2}{K^p}, \quad \lambda \propto \frac{NL^2}{e^K}.$$ (13)

## 5 Multiplicative Regularization

In this section, we analyze $E^{(m)}$ in (3) with linear multiplicative regularizer $f(k) = k$, and show that for ideal clusters it has a natural minimum at the correct number of clusters $K$. To do so, we show that for clusters with multiplicative regularizer $\Delta_{K-1,K}^{(m)} > 0$ while $\Delta_{K,K+1}^{(m)} < 0$:

$$\begin{cases} \Delta_{K-1,K}^{(m)} = (K-1)E_{K-1} - KE_K = \\ \quad \frac{1}{2}(K-1)VL^2 - KVR^2\alpha > 0, \text{for } d \geq 1, \ K \geq 2, \\ \\ \Delta_{K,K+1}^{(m)} = KE_K - (K+1)E_{K+1} = \\ \quad VR^2[\alpha - 2(K+1)\beta] < 0, \text{for } d \geq 2, \ K \geq 2. \end{cases}$$ (14)

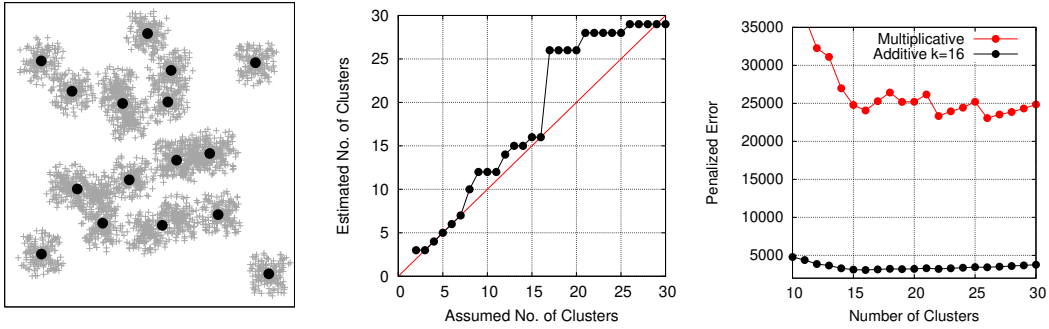

Figure 6: Clusters shown on the left are the same as in Fig. 5, but the inter-cluster distances are uniformly shrunk. The middle figure shows candidate solutions for additive regularizer are $K = 3, 4, 5, 6, 7, 16, 29$, while the figure on the right shows the local minima for multiplicative regularizer are at $K = 16, 19, 22, 26$. They agree only on $K = 16$, a rather reasonable solution shown on the left. Results were obtained by Algorithm 1.

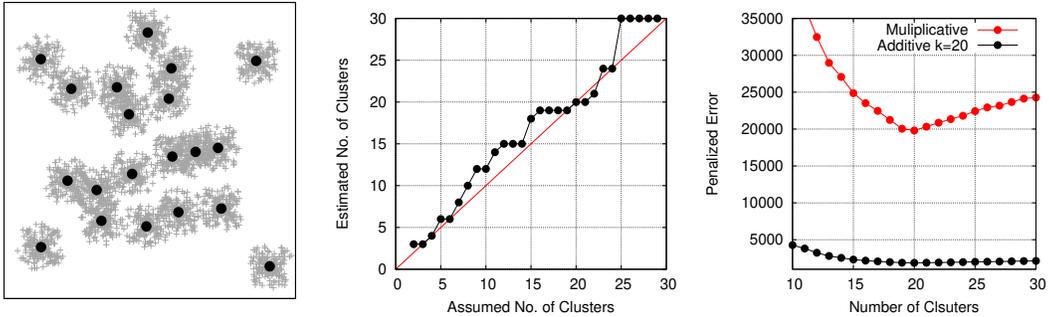

Figure 7: Clusters are the same as in Fig. 4, but the inter-cluster distances are uniformly shrunk. The middle figure shows candidate solutions for additive penalty are $K = 3, 4, 6, 7, 19, 20, 24$, while the figure on the right shows the minimum for multiplicative penalty is at $K = 20$, the correct number of clusters which both agree on shown on the left. These results were obtained by Algorithm 2.

The first inequality is satisfied since $L \geq 2R$ and $\alpha < 1$. The second inequality holds because $1 < \alpha/2\beta < 2$ for $d \geq 2$ (see Fig. 8 ). For the uninteresting $d = 1$ case (clustering on a straight line), for which $\alpha/2\beta = 4$, the inequality is satisfied when $K \geq 3$.

Note that the inequalities happen naturally; unlike in the additive regularization case where we must impose inequalities to ensure $E^{(a)}$ becomes minimum at $K$.

We also note that this formulation could be given a loose physics based interpretation. That is in $E^{(m)} = kE$, we may think of $E$ as the energy and $k$ as the inverse temperature; and as $k$ increases, or temperature decreases, the data points "crystalize" into proper clusters.

## 6 Experiments

In this section we present several experiments for selected number of dimensions and selected number of clusters. These include experiments where clusters deviate from the ideal cluster assumption to some degree in that they have significant overlaps or are not spherical, i.e., elongated. They show that multiplicative regularization has a stronger signature than additive regularization when clusters are ideal, and is generally more stable when the non-overlap assumption is violated.

To do the experiments we use both clustering Algorithms 1 and 2 presented in Sec. 3. For the additive regularization we use the procedure presented in Sec. 4.1, which yields candidate solutions when the *Assumed* and *Estimated* number of clusters are the same. We also compute the multiplicative regularization error $E_k^{(m)}$ as a function of $k$ from (3). Its minima indicate candidate solutions. In cases where there are ambiguities as to the correct number of clusters $K$, the agreement between solutions of additive and multiplicative regularization can resolve the ambiguity in some cases.

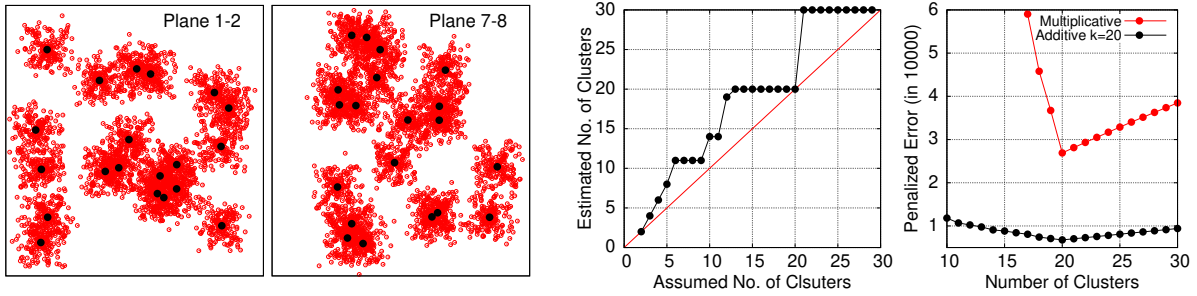

Figure 8: An example with 20 clusters in 8 dimensions. Left two figures are projections of clusters on two different 2D planes and create the impression they have significant overlap. The actual clusters are sufficiently well separated, hence both additive and multiplicative regularized k-means yield an unambiguous solution, as may be seen in the right two figures.

We note that the results of Algorithms 1 and 2 are identical when $k \geq K$, but often differ when $k < K$, even when clusters are ideal. This may be seen in Figures 4 and 5, and more noticeably in Figures 6 and 7. This is a well-known problem in k-means, as the resulting clusters are sensitive to initialization, and is also discussed in [13], in particular for well separated clusters. We also note that in cases where the clusters are not ideal other initializations may yield better solutions. However, we do not include those experiments here.

For all the experiments, in addition to the data and its computed cluster centroids, we present two figures. One is the "Estimated vs. Assumed No. of Clusters" for the aditive regularization case. When estimated and assumed number of clusters $k$ are the same, it indicates that $k$ is a candidate solution for additive regularization, as mentioned above. The second figure is "Penalized Error vs. Number of Clusters." In this figure (local) minima of the multiplicative regularization error indicate candidate solutions for the multiplicative penalty case. In this figure, for comparison, we also plot the additive regularization error for a selected number of clusters.

### 6.1 Simulated Data

Figures 2 and 3 are both 2D cases with $K = 10$ clusters each with approximately 100 points. From the "Estimated vs. Assumed" figure it can be seen that the additive penalty yields several candidate solutions, while multiplicative penalty yields only one solution at the correct number of clusters. It can also be seen that the additive penalized errors are significantly shallower than the multiplicative penalized error. Also note that as clusters overlap more, as in Fig. 3, both additive and multiplicative errors become shallower. The results shown in these figures are obtained with Algorithm 1. Algorithm 2 yields similar results.

Fig. 4 is a 2D case with $K = 20$ clusters each with about 200 data points. As can be seen there is little overlap among clusters, nevertheless additive penalty yields several candidate solutions shown in the "Estimated vs. Assumed Number of clusters" figure, while the multiplicative penalty yields only one solution at the correct value. We present the results of using Algorithm 1 in Fig. 4, and Algorithm 2 in Fig. 5. Note that when $k = 19$, the dumbbell shaped clusters in Figures 4 and 5 are different, while the results of the two algorithms are identical when $k \geq 20$.

In Figures 6 and 7 individual clusters are the same as in those of Figures 4 and 5, but the inter-cluster distances have been significantly reduced so that the clusters have substantial overlap. For this case, the results of Algorithms 1 and 2 differ substantially. While for Algorithm 1 the consensus of additive and multiplicative formulations is $K = 16$ (Fig. 6), for Algorithm 2 the consensus is the $K = 20$ clusters (Fig. 7).

Fig. 8 is an example with $K = 20$ clusters in $d = 8$ dimensions. The additive regularizer yields $k = 2, 20$ as candidate solutions, while the multiplicative regularizer yields only $k = 20$. We also plot $E_k^{(a)}$ computed with the $\lambda$ obtained for $K = 20$, and $E_k^{(m)}$, which show the comparative depth of their minima. Note that at higher dimensions clusters generally have less overlap, thus becoming more like ideal clusters with little overlap and unambiguous solutions. The results shown were obtained by Algorithm 1. Algorithm 2 yields similar solutions.

### 6.2 Textured Image

Fig. 9 is a mosaic of five Brodatz textures [36]. There are many alternative ways of representing the image for clustering. Here we show two alternatives–local image moments and local discrete cosine transform.

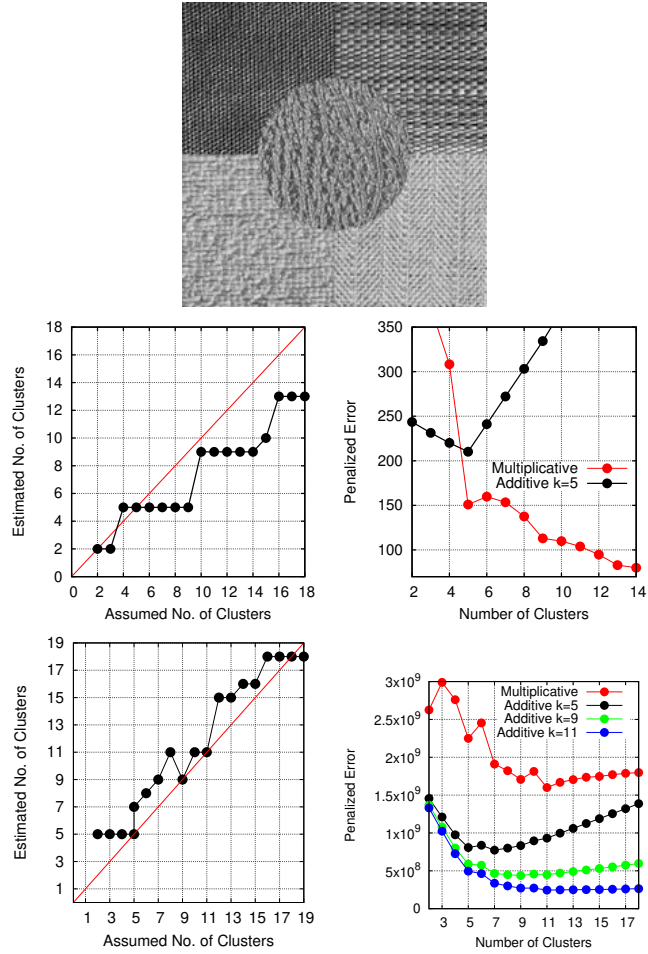

Figure 9: **Top row:** A composite image with 5 Brodatz textures. **Middle row:** The image is represented by 1,654 points in a $d = 6$ dimensional space obtained from local image moments. The combined k-means algorithms with additive and multiplicative penalties yield 5 clusters (textures). **Bottom row:** The image is represented by 1,430 points in a $d = 9$ dimensional space obtained from local discrete cosine transform. The combined k-means algorithms with additive and multiplicative penalties suggest $k = 5, 9, 11$ are candidate solutions.

First, we select 2,000 pixels at random and compute the first six moments in a $9 \times 9$ window. The pixels on the borders of the texture regions are sparse and appear like outliers. The clustering algorithms presented in Sec. 3 are sensitive to outliers, as are other clustering algorithms. Therefore, to deal with outliers we compute the point density around each of them, and eliminate those that have densities below a certain threshold. The remaining data is 1,654 points in a 6 dimensional space. In this figure we show the results of the penalized k-means algorithms. As may be seen, at $k = 5$ the additive penalty has a pronounced global minimum, while multiplicative penalty has a local minimum. The results shown are produced by Algorithm 1. Algorithm 2 produces similar results.

In the second alternative, we take 2,000 randomly selected $8 \times 8$ windows. Then we represent the windows by the first 9 coefficients of the discrete cosine transform. To remove the outliers (windows that straddle texture region boundaries), we perform the same culling procedure we mentioned above. As seen Fig. 9 (Bottom row), k-means with additive penalty indicate that $k = 5, 9, 11, 18$ are candidate solutions, while the multiplicative penalty indicates $k = 5, 9, 11$, the local minima of the penalized error, are candidate solutions. Thus, the combined k-means algorithms agree that that candidate solutions are $k = 5, 9, 11$.

### 6.3 Iris Data

Here we present the results for the well-known Anderson's Iris data with three species $K = 3$ and $N = 150$ data points in a 4-dimensional features space [3][12]. The data violates the ideal cluster assumptions in that all three clusters are elongated. The ratio of major to minor feature extensions, which should be 1 for ideal clusters, for this dataset are 4.2 for Iris Setosa, 2.6 for Versicolor, and 2.7 for Virginica. Moreover, there is significant overlap between two of the clusters. See Fig. 10. The results of Algorithm 1 and 2 are shown in Figures 11 and 12. For Algorithm 1 the consensus solution is $K = 3$ (Fig. 11). Even though consensus solution for Algorithm 2 is $K = 4$, additive penalty shows a strong preference for $K = 3$ as may be seen in the Penalized Error plot (Fig. 12).

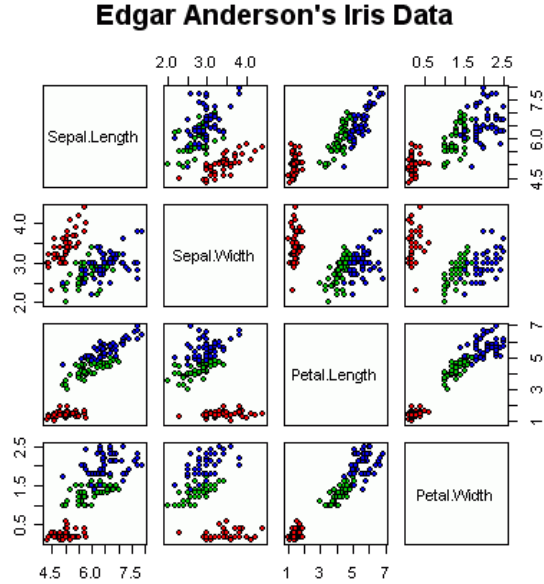

Figure 10: Plots of Iris data for all dimension pairs. Reproduced from [9] with permission.

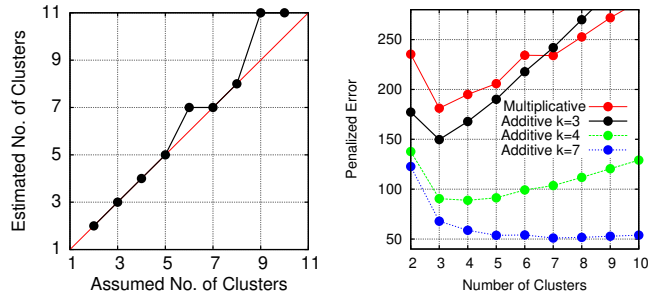

Figure 11: Results of Algorithm 1 for Iris data. (**Left**) Additive penalty yields $k = 2, 3, 4, 5, 7, 8$ as solution candidates. (**Right**) Multiplicative penalty has its minimum at $k = 3$, and a very shallow local minimum at $k = 7$. The consensus is $k = 3$ where both methods have a sharp minimum. Also plotted are penalized error as a function of $k$ for additive penalty with $\lambda$ estimated from $k = 3, 4, 7$.

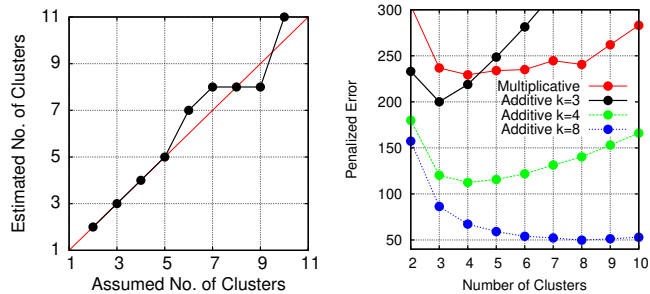

Figure 12: Results of Algorithm 2 for Iris data. (**Left**) Additive penalty yields $k = 2, 3, 4, 5, 8$ as solution candidates. (**Right**) Multiplicative penalty is rather ambiguous with its minimum at $k = 4$, and a local minimum at $k = 8$. While consensus is $k = 4$, additive penalty strongly indicates $k = 3$ as the solution. Also plotted are errors as a function of $k$ for additive penalty with $\lambda$ estimated from $k = 3, 4, 8$.

## 7    Concluding Remarks

Penalized k-means algorithms are frequently used to estimate the correct number of distinct clusters in a dataset. In this paper, we discussed the regularizd (penalized) k-means algorithm with both additive and multiplicative penalties. For the additive regularization, which is more commonly used in practice, an open problem has been: what is a good value for the coefficient of the penalty term? To address this important question, we derived rigorous upper and lower bounds for the coefficient of the additive penalty term. The bounds are derived for ideal clusters (same-size non-overlapping spheres), which have stronger underlying assumptions than those of k-means clusters (same-size spherically symmetric clusters) and provide useful guidelines for selecting the value of the penalty coefficient. In practice, regularized k-means algorithm is often used beyond its valid range. Namely, in cases where clusters have different sizes, are not spherically symmetric but elongated, or have significant overlaps.

We have empirically investigated the performance of regularized k-means algorithms as clusters violate the ideal assumption. In particular, as cluster overlaps grow, when clusters are not spherical, and when the dataset contains outliers. Based on experiments, our observations are (a) when clusters are ideal, k-means algorithm with multiplicative penalty generally produces stronger signatures for the correct number of clusters than k-means with additive penalty. (b) When there are outlier data points, it appears that k-means with additive penalty yields signatures that are more reliable. (c) As the level of overlap among clusters increases, k-means algorithm with both additive and multiplicative penalties may suggest multiple solutions. For cases where the solution is ambiguous, we presented a procedure that looks for agreement between the two sets of suggested solutions, which could potentially resolve or reduce the ambiguity. What the limits of this disambiguation procedure are needs further theoretical analysis or more exhaustive empirical evidence. We also note that for the sake of consistency with the ideal cluster assumption, the experimental results presented were obtained using Algorithms 1 or 2 or both. Other initializations may yield different or better results.

The procedure we presented in this paper is particularly useful for high-dimensional problems, where projection to two or three dimensions (by such algorithms as Sammon mapping [33]) for visual inspection are not possible.

The results presented in this paper may be further extended in a number of directions that relax the ideal cluster assumptions. These include (a) rigorous derivation of the limits of non-overlap assumption; (b) extend to spherically symmetric normal distribution of data points rather than spherical distributions, which would allow greater tolerance of overlapping clusters; (c) relax the assumption that the number of points in each cluster be approximately the same; (d) theoretical analysis and the algorithm for finding the best clustering is sensitive to outliers, develop more sophisticated methods for detecting and discarding outliers.

## Acknowledgement

This work was supported by the Office of Naval Research award N0001425GI01276.

# References

[1] S. Ahmadian, A. Norouzi-Fard, O. Svensson, J. Ward. Better Guarantees for k-Means and Euclidean k-Median by Primal-Dual Algorithms. *Proc. FOCS*, pp. 61-72, 2017.

[2] R.C. de Amorim and C. Henning. Recovering the number of clusters in data sets with noise features using feature rescaling factors. arXiv:1602.06989v1, Feb. 2016.

[3] E. Anderson. The species problem in Iris. *Annals of the Missouri Botanical Garden*, 23(3):457–509, 1936.

[4] D. Arthur and S. Vassilvitskii. k-means++: The Advantages of Careful Seeding. *Proc. Annual ACM-SIAM Symposium on Discrete Algorithms (SODA)*, pp. 1027-1035, 2007.

[5] O. Bachem, M. Lucic, S.H. Hassani, and A. Krause. Fast and Provably Good Seedings for k-Means. *Proc. Neural Information Processing Systems (NIPS)*, pp. 55-63, 2016.

[6] C. Baldassi. Recombinator-k-means: Enhancing k-means++ by seeding from pools of previous runs. arXiv:1905.00531v1, May 2019

[7] A. Blum, J. Hopcroft, and R. Kannan. *Foundations of Data Science*, Cambridge University Press, 2020.

[8] H. Bozdogan. Model selection and Akaike's information criterion (AIC): the general theory and its analytical extensions. Psychometrika, 52(3):345–370, 1987.

[9] P.J.A. Cock. https://warwick.ac.uk/fac/sci/moac/people/students /peter_cock/r/iris_plots/.

[10] R.O. Duda, P. E. Hart, and D.G. Stork. *Pattern Classification*. Wiley, 2003.

[11] A. Fischer. On the number of groups in clustering. *Statistics and Probability Letters*, 81:1771–1781, 2011.

[12] A. Frank and A. Asuncion. UCI Machine Learning Repository. University of California, Irvine, School of Information and Computer Sciences, 2010. http://archive.ics.uci.edu/ml.

[13] P. Franti and S. Sieranoja. How much can k-means be improved by using better initialization and repeats? *Pattern Recognition*, 93(9):95-112, 2019.

[14] W. Fu and P.O. Perry. Estimating the number of clusters using cross-validation. arXiv:1702.02658v1, Feb. 2017.

[15] A. Fujita, D.Y. Takahashi, and A.G. Patriota. A non-parametric method to estimate the number of clusters. *Computational Statistics and Data Analysis*, 73:27-39, 2014.

[16] A. Gersho and R.M. Gray. *Vector Quantization and Signal Compression*. Kluwer Academic Publishers, 1992.

[17] I.S. Gradshteyn and I.M. Ryzhik. *Table of Integrals, Series, and Products*, Academic Press, 4th edition, p. 369, 1965.

[18] A.K. Jain and R.C. Dubes. *Algorithms for Clustering Data*. Prentice-Hall, 1988.

[19] A.K. Jain. Data clustering: 50 years beyond K-means. *Pattern Recognition Letters*, 31(8):651-666, 2010.

[20] B. Kamgar-Parsi, J.A. Gualtieri, J.E. Devaney, and B. Kamgar-Parsi. Clustering with neural networks. *Biological Cybernetics*, 63(3):201-208, 1990.

[21] B. Kamgar-Parsi and B. Kamgar-Parsi. Dynamical Stability and Parameter Selection in Neural Optimization. *Proc. International Joint Conference on Neural Networks,* 4:566-571, 1992.

[22] B. Kamgar-Parsi and B. Kamgar-Parsi. Penalized K-Means Algorithms for Finding the Number of Clusters. *Proc. 2020 International Conference on Pattern Recognition (ICPR 2020)*, Milan, Italy, Jan. 10-15, 2021, pp. 969-974.

[23] T. Kanungo, D.M. Mount, N.S. Netanyahu, C.D. Piatko, R. Silverman, and A.Y. Wu. An Efficient k-Means Clustering Algorithm: Analysis and Implementation. *IEEE Trans. Pattern Analysis and Machine Intelligence*, 24(7):881-892, 2002.

[24] W.J. Krzanowski and Y.T. Lai. A criterion for determining the number of clusters in a data set. *Biometrics*, 44:23–34, 1988.

[25] B. Kulis and M.I. Jordan. Revisiting k-means: New Algorithms via Bayesian Nonparametrics. *Proc. International Conference on Machine Learning (ICML)*, Edinburgh, UK, 2012.

[26] W. Li, E.K. Ryu, S. Osher, W. Yin, and W. Gangbo. A parallel method for earth mover's distance. *UCLA Comput. Appl. Math. Pub. (CAM) Rep.*, 17-12, 2017.

[27] Y. Linde, A. Buzo, and R. Gray. An Algorithm for Vector Quantizer Design. *IEEE Trans. Communications*, 28:84-95, 1980.

[28] S. P. Lloyd. Least squares quantization in PCM. *IEEE Trans. Information Theory*, 28(2):129-137, 1982. Appeared first in a 1956 AT&T Bell Labs memorandum.

[29] C.D. Manning, P. Raghavan, and H. Schutze. *Introduction to Information retrieval*. Cambridge University Press, 2008.

[30] K.P. Murphy. *Machine Learning: A Probabilistic Perspective*. MIT Press, 2012.

[31] K. Rose, E. Gurewitz, and G.C. Fox. Vector quantization by deterministic annealing. *IEEE Trans. Information Theory*, 38(4):1249-1257, 1992

[32] P.J. Rousseeuw. Silhouettes: A graphical aid to the interpretation and validation of cluster analysis. *Journal of Computational and Applied Mathematics,* 20:53-65, 1987.

[33] J.W. Sammon. A Nonlinear Mapping for Data Structure Analysis. *IEEE Trans. Computers*, 18(5):401–409, 1969.

[34] G. Schwarz. Estimating the dimension of a model. *The Annals of Statistics*, 6(2):461–464, 1978.

[35] C.A. Sugar and G.M. James. Finding the number of clusters in a dataset: an information-theoretic approach. *Journal of the American Statistical Association*, 98(463):750-763, 2003.

[36] P. Szczypiński, M. Strzelecki, and A. Materka. MaZda A Software for Texture Analysis. Proc. Int'l Symp. on Information Technology Convergence, ISITC 2007.

[37] R. Tibshirani, G. Walther, and T. Hastie. Estimating the number of clusters in a data set via the gap statistic. *Journal of the Royal Statistical Society*, 63(2):411-423, 2001.

[38] R. Vidal, Y. Ma, and S. Sastry. *Generalized Principal Component Analysis*, Springer-Verlag, 2016.

[39] https://en.wikipedia.org/wiki/Determining_the_number_of_ clusters_in_a_data_set.

## Appendix A: Single-Cluster Variances

In this Appendix, we calculate the volume of a sphere in the $n$-dimensional Euclidean space, as well as the within-cluster variances for the sphere, and the centroid and the within-cluster variances for the half-sphere, the dumbbell, and the uneven dumbbell. Note that while in the body of the paper we denote the dimension by $d$, in this appendix we denote it by $n$ to avoid confusion with the differential element $d$ in the integrals.

### A.1 Sphere

The volume element in the $n$-dimensional spherical coordinate system (in terms of radius $r$ and angles $\phi_1, \ldots, \phi_{n-1}$) is:

$$dV = [dr\, r^{n-1}]\, [d\phi_1\, (\sin \phi_1)^{n-2}]\, [d\phi_2\, (\sin \phi_2)^{n-3}] \cdots [d\phi_{n-2}\, (\sin \phi_{n-2})]\, [d\phi_{n-1}]. \tag{A-1}$$

Hence the volume of a sphere of radius $R$ may be calculated from:

$$V = [\int_0^R dr\, r^{n-1}]\, [2 \int_0^{\pi/2} d\phi_1\, (\sin \phi_1)^{n-2}] \cdots [2 \int_0^{\pi/2} d\phi_{n-2}\, (\sin \phi_{n-2})]\, [4 \int_0^{\pi/2} d\phi_{n-1}]. \tag{A-2}$$

The integrals over spherical angles $\phi_1 \cdots \phi_{n-1}$ can be calculated from the following formula [17],

$$\int_0^{\pi/2} d\phi\, (\sin \phi)^n = \frac{\Gamma(\frac{n+1}{2})\Gamma(\frac{1}{2})}{2\,\Gamma(\frac{n+2}{2})}, \quad \text{for all integers } n \geq 0, \tag{A-3}$$

where $\Gamma(z+1) = z\Gamma(z)$, the generalized factorial function, is:

$$\text{For integer } n: \ \Gamma(n) = (n-1)!, \ \Gamma(n + \frac{1}{2}) = \frac{(2n)!\sqrt{\pi}}{4^n\, n!}. \tag{A-4}$$

With repeated use of (A-3), Eq. (A-2) simplifies to:

$$V = \frac{\pi^{\frac{n}{2}}}{\Gamma(\frac{n+2}{2})}\, R^n. \tag{A-5}$$

The within-cluster variance, or clustering error, for the $n$-dimensional sphere is given by:

$$E_s = \int_{sphere} dV \parallel \mathbf{x} - \mathbf{c} \parallel^2, \tag{A-6}$$

and may be expressed in the spherical coordinate system (the centroid $\mathbf{c}$ is at the origin),

$$E_s = [\int_0^R dr\, r^{n+1}]\, [2 \int_0^{\pi/2} d\phi_1\, (\sin \phi_1)^{n-2}] \cdots [2 \int_0^{\pi/2} d\phi_{n-2}\, (\sin \phi_{n-2})]\, [4 \int_0^{\pi/2} d\phi_{n-1}]. \tag{A-7}$$

This can be similarly simplified to:

$$E_s = V\, R^2\, \alpha, \quad \text{with } \alpha = \frac{n}{n+2}. \tag{A-8}$$

### A.2 Half-sphere

The centroid $\mathbf{c}$ of the half-sphere is on the axis that is orthogonal to the equator, say axis 1, that is $\mathbf{c} = (\rho, 0, 0, \ldots, 0)$. Its distance, $\rho$, from the equator may be calculated from

$$\tfrac{1}{2}\, V\, \rho = \int_{hs} dV\, r = [\int_0^R dr\, r^n][\int_0^{\pi/2} d\phi_1 (\sin \phi_1)^{n-2} \cos \phi_1]$$
$$[2 \int_0^{\pi/2} d\phi_2\, (\sin \phi_2)^{n-3}] \cdots [2 \int_0^{\pi/2} d\phi_{n-2}\, (\sin \phi_{n-2})]\ [4 \int_0^{\pi/2} d\phi_{n-1}]. \tag{A-9}$$

Subscript $hs$ indicates integral over the volume of the half-sphere. This leads to

$$\rho = R\gamma, \quad \text{with } \gamma = \frac{\Gamma(\frac{n+2}{2})}{\sqrt{\pi}\,\Gamma(\frac{n+3}{2})}. \tag{A-10}$$

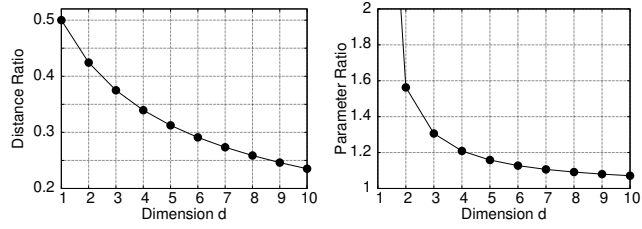

Figure 13: (**Left**) The ratio $\gamma = \rho/R$ as a function of dimension $d$, where $R$ is the sphere radius and $\rho$ is distance of the half-sphere centroid from its equator plane; $\rho \to 0$ as $d \to \infty$. (**Right**) The ratio $\alpha/2\beta$ as a function of dimension $d$. The ratio goes to 1 as $d \to \infty$.

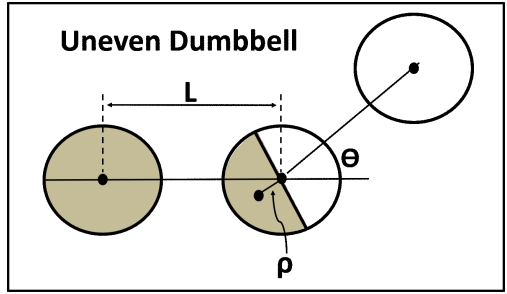

Figure 14: A sketch of an uneven dumbbell.

The within-cluster variance for the half-sphere is given by:

$$E_h = \int_{half-sphere} dV \parallel \mathbf{r} - \mathbf{c} \parallel^2, \tag{A-11}$$

which reduces to the following,

$$E_h = \tfrac{1}{2} E_s - \tfrac{1}{2} V \rho^2 = V R^2 \beta, \tag{A-12}$$

$$\text{with } \beta = \tfrac{1}{2} (\tfrac{n}{n+2} - \gamma^2) = \tfrac{1}{2}(\alpha - \gamma^2).$$

### A.3 Dumbbell

The within-cluster variance for the dumbbell with two equal size spheres may be expressed as,

$$E_d = 2E_s + 2V(\frac{L}{2})^2 = 2VR^2\alpha + \frac{1}{2}VL^2, \tag{A-13}$$

where $L$ is the distance between the centers of the two spheres forming the dumbbell.

A quantity of interest is the ratio $\dfrac{\alpha}{2\beta}$, which ranges between 1 and 4. For $n = 1$ (straight line) we have $\dfrac{\alpha}{2\beta} = 4$, and as $n \to \infty$ it approaches 1, thus $1 < \dfrac{\alpha}{2\beta} \le 4$. See Fig. 13.

### A.4 Uneven Dumbbell

The shaded areas of Fig. 14 show an uneven dumbbell. It is straight forward to show that the error $E_u$ for this uneven dumbbell is given by,

$$E_u = E_s + E_h + \frac{V}{3}[(L - \rho\cos\theta)^2 + (L - \rho\sin\theta)^2]. \tag{A-14}$$

At $\theta = 0$ or $\frac{\pi}{2}$ the error $E_u$ has its smallest value. Since $E_u$ is related to the upper bound of $\lambda$ and we want to derive the tightest bounds, we use its minimum value given by,

$$E_u = E_s + E_h + \frac{V}{3}(L - \rho)^2 = VR^2\alpha + VR^2\beta + \frac{V}{3}(L - \rho)^2. \tag{A-15}$$

In the above derivation, we assumed that the two outer spheres are at the same distance $L$ from the center sphere. If we perturb the 3-sphere configuration such that one remains at the distance of $L$ from the center sphere and the other moves away to the distance of $L+\epsilon$, where $\epsilon \ll L$, then the composite error for the two uneven dumbbells increases as $O(\epsilon^2)$. This increase will continue until the sphere is at a distance of $L+R$ from the center sphere, at which point the two uneven dumbbells transition to a configuration with a sphere and a perfect dumbbell. As we noted above, since we are interested in deriving the tightest upper bound for $\lambda$, we will consider the equal-distance configuration with $\theta = 0$.

## Appendix B: Resolving Upper Bound Ambiguity for $\lambda$

In this Appendix, we resolve the ambiguity regarding the upper bound for $\lambda$ in the penalized k-means with linear additive penalty and show that the dumbbell configuration yields the tighter bound.

Let us denote the clustering error (1) for $k$ clusters by $E_k$. Also let $K$ be the correct number of clusters. In Sec. 3, we see that the lower bound for $\lambda$ is obtained from,

$$\lambda > E_K - E_{K+1} = V\rho^2. \tag{B-1}$$

This lower bound is unique for ideal clusters since both $E_K$ and $E_{K+1}$ are unique. Similarly the upper bound for $\lambda$ is given by,

$$\lambda < E_{K-1} - E_K, \tag{B-2}$$

which cannot be determined uniquely. Because while $E_K$ is unique, i.e. composed of $K$ sphere, $E_{K-1}$ is not as it consists of spheres along with either a perfect dumbbell or two uneven dumbbells. Thus, we derive the lower bound for the case when $K-1$ clusters are composed of (a) one perfect dumbbell plus $K-2$ spheres, and (b) two uneven dumbbells and $K-3$ spheres. We seek the case that is more general and yields the tightest bound.

For the case with one perfect dumbbell and $K-2$ spheres, we have (subscripts and superscripts $d$ and $u$ refer to dumbbell and uneven dumbbell, respectively),

$$E_{K-1}^{(d)} = (K-2)E_s + E_d. \tag{B-3}$$

Noting that $E_K = KE_s$, (B-2) becomes,

$$\lambda < \lambda_d = \frac{1}{2}VL^2. \tag{B-4}$$

For the case with two uneven dumbbells and $K-3$ spheres we have,

$$E_{K-1}^{(u)} = (K-3)E_s + 2E_u, \tag{B-5}$$

and (B-2) reduces to,

$$\lambda < \lambda_u = \frac{1}{3}V(2L^2 - 4LR\gamma - R^2\gamma^2). \tag{B-6}$$

The right hand side is always positive and is at least $5VR^2/4$. This is because $L \geq 2R$ for non-overlapping spheres, and $\gamma = 0.5$ for dimension $n = 1$ and decreases as the dimension increases becoming zero as $n \to \infty$.

Since we are interested in the tightest bounds, we examine which of the two upper bounds, $\lambda_d$ or $\lambda_u$, is smaller. To do so, we examine the conditions under which the inequality $\lambda_u < \lambda_d$ is satisfied. Combining Eqns (B-4) and (B-5), this inequality may be written as,

$$2\gamma^2 + 8(\frac{L}{R})\gamma - (\frac{L}{R})^2 > 0. \tag{B-7}$$

As we have noted above $L \geq 2R$. For $L = 2R$ the inequality is satisfied if $\gamma > 0.243$, that is when dimension $n \leq 9$. For $L = 3R$ the inequality is satisfied if $\gamma > 0.364$, that is when $n \leq 3$. For $L = 4R$ the inequality is satisfied if $\gamma > 0.485$, that is only when $n = 1$. And for $L = 5R$ the inequality is satisfied if $\gamma > 0.607$, which cannot be realized for any dimension. Moreover, $\lambda_u \to 2VL^2/3$ as the dimension $n \to \infty$ (or $\gamma \to 0$), which is larger than $\lambda_d$. Thus, $\lambda_d$ is a better upper limit because it is independent of the dimension, and is also smaller than $\lambda_u$ for most dimensions. Hence, when $k = K-1$ the clustering with one perfect dumbbell and $K-2$ spheres yields a better upper limit.

## Appendix C: Bounds for $\lambda$

In this Appendix we derive bounds for the coefficient $\lambda$ in the additive penalized k-means when the penalty function has polynomial, logarithmic, and exponential dependence on the number of clusters $K$. The additive penalty, Eq. (2), to have a minimum at $K$ clusters, must satisfy the following inequalities:

$$
\begin{cases}
\Delta_{K-1,K}^{(a)} = E_{K-1} - E_K \\
\quad = \frac{1}{2}VL^2 - \lambda(f(K) - f(K-1)) > 0, \ \text{ for all } \ n \geq 1 \ \text{ and } \ K \geq 1, \\
\\
\Delta_{K,K+1}^{(a)} = E_K - E_{K+1} \\
\quad = VR^2(\alpha - 2\beta) - \lambda(f(K+1) - f(K)) < 0, \ \text{ for all } \ n \geq 1 \ \text{ and } \ K \geq 1.
\end{cases}
\tag{C-1}
$$

The assumption is that when there are $K-1$ clusters, one of the clusters is a perfect dumbbell while the other $K-2$ clusters are spheres, as we argued in Appendix B. Also, as before $L$ is the distance between the centers of the two spheres forming the dumbbell. Noting that $R^2(\alpha - 2\beta) = \rho^2$, the two inequalities in (C-1) maybe combined and written as:

$$
\frac{V\rho^2}{f(K+1) - f(K)} < \lambda < \frac{VL^2}{2[f(K) - f(K-1)]}.
\tag{C-2}
$$

To estimate the value of $\lambda$, we remember that the volume $V$ is the average number of data points in each cluster, i.e., $V \approx N/K$, $\rho$ is given by (A-10) and is less than $R/2$ and goes to zero as the dimension $n$ increases.

We consider three functional forms for $f(K) = \ln K, K^p, e^K$. For these penalty terms (C-2) becomes,

$$
\begin{cases}
f(K) = \ln K : \ N\rho^2[1 + \frac{1}{K} + O(\frac{1}{K^2})] < \lambda < \frac{1}{2}NL^2[1 + \frac{1}{K} + O(\frac{1}{K^2})] \\
\\
f(K) = K^p : \ \frac{N\rho^2}{pK^p}[1 - \frac{p-1}{2K} + O(\frac{p^2}{K^2})] < \lambda < \frac{NL^2}{2pK^p}[1 + \frac{p-1}{2K} + O(\frac{p^2}{K^2})] \\
\\
f(K) = e^K : \ \frac{N\rho^2}{(e-1)e^K} < \lambda < \frac{NL^2}{2(1-\frac{1}{e})e^K}
\end{cases}
\tag{C-3}
$$

where we assume $K$ is sufficiently greater than $p$ for the polynomial case. It can be seen that $\lambda$ depends on the correct number of clusters $K$, which is not known a priori. In particular, for the commonly used linear penalty function, $f(K) = K$, we get:

$$
\frac{N\rho^2}{K} < \lambda < \frac{NL^2}{2K}.
\tag{C-4}
$$